\documentclass[letterpaper]{article} % DO NOT CHANGE THIS
\usepackage[preprint]{aaai24}  % DO NOT CHANGE THIS
\usepackage{times}  % DO NOT CHANGE THIS
\usepackage{helvet}  % DO NOT CHANGE THIS
\usepackage{courier}  % DO NOT CHANGE THIS
\usepackage[hyphens]{url}  % DO NOT CHANGE THIS
\usepackage{graphicx} % DO NOT CHANGE THIS
\usepackage{natbib}  % DO NOT CHANGE THIS AND DO NOT ADD ANY OPTIONS TO IT
\usepackage{caption} % DO NOT CHANGE THIS AND DO NOT ADD ANY OPTIONS TO IT
\usepackage{algorithm}
\usepackage{algorithmic}

\usepackage{newfloat}
\usepackage{listings}
\DeclareCaptionStyle{ruled}{labelfont=normalfont,labelsep=colon,strut=off} % DO NOT CHANGE THIS
\floatstyle{ruled}
\newfloat{listing}{tb}{lst}{}
\floatname{listing}{Listing}
\usepackage{siunitx}
\usepackage{amsmath}
\usepackage{amsthm}
\usepackage{amsfonts}
\usepackage{booktabs}
\usepackage{subcaption}

\newcommand{\dummyhref}[2]{{#2}}

\title{Learn to Follow: Decentralized Lifelong Multi-agent Pathfinding via Planning and Learning}
\author {
    Alexey Skrynnik\textsuperscript{\rm 1,2}, 
    Anton Andreychuk\textsuperscript{\rm 1},
    Maria Nesterova\textsuperscript{\rm 2}, \\
    Konstantin Yakovlev\textsuperscript{\rm 2,1},
    Aleksandr Panov\textsuperscript{\rm 1,2}
}
\affiliations {
    \textsuperscript{\rm 1}AIRI, Moscow, Russia\\
    \textsuperscript{\rm 2} Federal Research Center for Computer Science and Control of Russian Academy of Sciences, Moscow, Russia\\
    
    Corresponding author\thanks{Preprint. Under review}:  \dummyhref{mailto:skrynnikalexey@gmail.com}{skrynnikalexey@gmail.com}
}

\usepackage{bibentry}
\begin{document}

\maketitle

\begin{abstract}
    Multi-agent Pathfinding (MAPF) problem generally asks to find a set of conflict-free paths for a set of agents confined to a graph and is typically solved in a centralized fashion. 
    %Typically, the graph and the agents' start and goal locations are known in advance and a centralized planning algorithm is utilized to generate a solution. 
    Conversely, in this work, we investigate the decentralized MAPF setting, when the central controller that posses all the information on the agents' locations and goals is absent and the agents have to sequientially decide the actions on their own without having access to a full state of the environment. We focus on the practically important lifelong variant of MAPF, which involves continuously assigning new goals to the agents upon arrival to the previous ones. To address this complex problem, we propose a method that integrates two complementary approaches: planning with heuristic search and reinforcement learning through policy optimization. Planning is utilized to construct and re-plan individual paths. We enhance our planning algorithm with a dedicated technique tailored to avoid congestion and increase the throughput of the system. We employ reinforcement learning to discover the collision avoidance policies that effectively guide the agents along the paths. The policy is implemented as a neural network and is effectively trained without any reward-shaping or external guidance.
    We evaluate our method on a wide range of setups comparing it to the state-of-the-art solvers. The results show that our method consistently outperforms the learnable competitors, showing higher throughput and better ability to generalize to the maps that were unseen at the training stage. Moreover our solver outperforms a rule-based one in terms of throughput and is an order of magnitude faster than a state-of-the-art search-based solver. 
\end{abstract}

\section{Introduction}
\label{sec:intro}
Multi-agent pathfinding (MAPF)~\cite{stern2019multi} is a challenging problem that has been getting increasing attention recently. It is often studied in the AI community with the following assumptions. The agents are confined to a graph, and at each timestep, an agent can either move to an adjacent vertex or stay at the current one. A central controller possesses information about the graph and the agents' start and goal locations. This unit is in charge of constructing a set of conflict-free plans for all the agents. Thus, a typical setting for MAPF can be attributed as \emph{centralized} and \emph{fully observable}.

\begin{figure}[t!]
    \centering
    \includegraphics[width=1.0\linewidth]{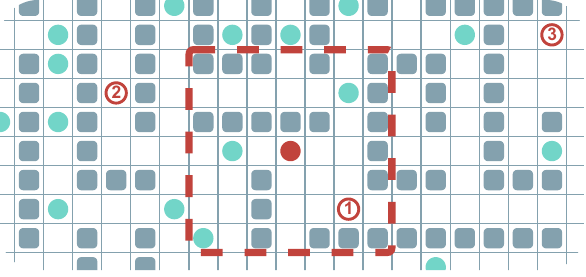}
    \vspace{-10px}
    \caption{An example of a decentralized LMAPF instance. Agents are depicted as filled circles. The dashed line illustrates the red agent's ego-centric field-of-view, where the other observed agents are colored in blue. The red circles with numbers represent the goals that the agent needs to reach. The next goal is only revealed when the current one is achieved. }
    \label{fig:visual-abstract}
    \vspace{-10px}
\end{figure}

In many real-world domains, however, it is not possible, from the engineering perspective, to design such a central controller that has a stable connection to all the agents (robots) and obtains a full knowledge of the environment all the time. For example, consider a fleet of service robots delivering some items in a human-shared environment, e.g., the robots delivering medicine in the hospital. Each of these robots is likely to have access to the global map of the environment (e.g., the floor plan), possibly refined through the robot's sensors. However, the connection to the central controller may not be consistent. Thus, the latter may not have accurate data on the robots' locations and, consequently, cannot provide valid MAPF solutions. In such scenarios, \emph{decentralized approaches} to the MAPF problems, when the robots themselves have to decide their future paths based on their local observations, as depicted in Fig.~\ref{fig:visual-abstract}, are essential. 
In this work, we aim to develop such an efficient decentralized approach.

It is natural to frame the decentralized Multi-Agent Path Finding (MAPF) problem as a sequential decision-making problem where, at each timestep, each agent must choose and execute an action that will advance it toward its goal while ensuring that other agents can also reach their goals. The result of solving this problem is a policy that, at each moment, specifies which action to execute. To form such a policy, learnable methods are commonly used nowadays, such as reinforcement learning (RL), which proves particularly beneficial in tasks with incomplete information~\cite{Mnih2015,Rashid2018a,Hafner2020a}.
However, even state-of-the-art RL methods generally struggle with solving long-horizon problems with the involved casual structure~\cite{Milani2020,hafner2023mastering}, and they are often inferior to the search-based, planning methods when solving problems with hard combinatorial structure~\cite{kansky2023pushworld}.

Indeed, numerous learnable methods tailored to MAPF settings are already known, such as PRIMAL~\cite{sartoretti2019primal}, PRIMAL2~\cite{damani2021primal}, DHC~\cite{ma2021distributed}, PICO~\cite{Li2022MultiAgentPF}, SCRIMP~\cite{wang2023scrimp} to name a few. These methods either rely on the complex training procedures that typically involve manual reward-shaping, external demonstrations etc., or on the communication (data sharing) between the agents. Moreover these methods often do not generalize well, i.e. their performance degrades significantly when they solve problem instances on the maps that are not alike the ones used for training. 

%The additional challenges that make the MAPF problems demanding for RL are as follows. First, we want the policy to be highly generalizable to the unseen environments, that may differ significantly in scale and topology from the ones used for training. Second, MAPF problems are naturally dependent on the goal locations of the agents, meaning that even in the same environment (map), the goals may vary significantly. Finally, effectively training in a complex observation and action spaces poses challenges even for state-of-the-art multi-agent reinforcement learning (MARL) methods.

To this end, the current paper suggests that the MAPF problem should not be solved directly by RL but rather in combination and vivid interaction with the heuristic search algorithm. This idea is put into practice via the following pipeline. 
%decomposed into a series of sub-tasks utilizing heuristic search algorithms in order to then solve these sub-tasks efficiently with a learnable policy. 
%The general pipeline of our solution is the following. 
Each agent plans an individual path to its goal by a heuristic search algorithm without taking the other agents into account. Moreover, an additional technique is introduced for planning that is dedicated specifically to dispersing the agents over the workspace via penalizing the paths that are likely to cause deadlocks. Upon path construction, a learnable policy is invoked to reach the first waypoint of this path. This policy is obtained through a decentralized training and is aimed at following the path on the one hand, while making necessarily detours to avoid collisions and let the other agents progress toward their goals on the other hand.

%a waypoint on this path is chosen in some vicinity of the agent, which becomes its local goal. To reach it, a learnable policy is utilized, which takes both static obstacles and locally observable agents into account. 
%Once a waypoint is reached or the agent goes too far away from it the cycle repeats.

Empirically, we compare our method, which we name \textsc{Follower}, to a range of both learnable and non-learnable state-of-the-art competitors and show that it \emph{i}) consistently outperforms the learnable competitors in terms of solution quality; \emph{ii}) better generalizes to the unseen environments compared to the other learnable solvers; \emph{iii}) outperforms a state-of-the-art rule-based centralized solver in terms of solution quality; \emph{iv}) scales much better to the large numbers of agents in terms of computation time compared to the state-of-the-art search-based centralized solver.

\section{Related Works}
\label{sec:related}

\paragraph{Lifelong MAPF} LMAPF is an extension of MAPF when the agents are assigned new goals upon reaching their current ones. Similarly, in (online) multi-agent pickup and delivery (MAPD), agents are continuously assigned tasks comprising two locations that the agent has to visit in a strict order: pickup location and delivery location. Typically, the assignment problem is not considered in LMAPF/MAPD. However, some works also consider task assignment, such as~\cite{liu2019task,chen2021integrated}.

\citeauthor{ma2017lifelong}~\shortcite{ma2017lifelong} propose several variants to tackle MAPD differing in the amount of data the agents share. Yet, even the decoupled (as attributed by the authors) algorithms based on Token Swapping rely on global information, i.e., the one provided by the central unit. An enhanced Token Swapping variant that considers kinematic constraints was introduced in~\cite{ma2019lifelong}. In~\cite{okumura2019priority} an efficient rule-based re-planning approach to solve MAPF that is naturally capable of solving LMAPF/MAPD problems is introduced -- PIBT. It does not rely on the several restrictive assumptions of Token Swapping and is empirically shown to outperform the latter. We compare with PIBT and demonstrate that our method is better regarding solution quality.

Finally, one of the most recent and effective LMAPF solvers is the RHCR algorithm presented in~\cite{li2021lifelong}. It draws upon the idea of bounded planning, i.e., constructing not a complete plan but rather its initial part. RHCR is a centralized solver that relies on the full knowledge of the agents' locations, current paths, goals, etc. In this work, we empirically compare with RHCR and show that our method scales much better to many agents in computation time.

\paragraph{Decentralized MAPF} This setting entails that the paths/actions of the agents are not decided by a central unit but by the agents themselves. Numerous approaches, especially the ones tailored to the robotics applications, boil this problem down to reactive control~\cite{lumelsky1997decentralized,van2008reciprocal,zhu2022decentralized}. These methods, however, are often prone to deadlocks. Several MAPF algorithms can also be implemented in a decentralized manner. For example, ~\citeauthor{wang2011mapp}~\shortcite{wang2011mapp} introduce MAPP algorithm that relies on individual pathfinding for each agent and a set of rules to determine priorities and choose actions to avoid conflicts when they occur along the paths.
% ~\cite{okumura2019priority} explores a PIBT algorithm in which the agents also pick their actions individually (at each timestep) based on specific rules. 
In general, most rule-based MAPF solvers, like the previously mentioned PIBT~\cite{okumura2019priority}, or another seminal MAPF solver Push And Rotate~\cite{de2013push}, can be implemented in such a way that each agent locally decides its actions. However, in this case, the implicit assumption is that the agents can communicate to share relevant information (or that they have access to the global MAPF-related data). By contrast, our work assumes that the agents cannot reliably communicate with each other or a central unit, which significantly increases the complexity of the problem.

\paragraph{Learnable MAPF} This direction has recently received increased attention. In ~\cite{sartoretti2019primal}, a seminal PRIMAL method was introduced. It utilizes reinforcement learning and imitation learning to solve MAPF in a decentralized fashion. Later in~\cite{damani2021primal}, it was enhanced and tailored explicitly to LMAPF. The new version was named PRIMAL2. Since numerous learning-based MAPF solvers have emerged, it has become common to compare against PRIMAL/PRIMAL2 (we also compare with it in our work). For example, \citeauthor{riviere2020glas}~\shortcite{riviere2020glas} propose another learning-based approach tailored explicitly to agents with a non-trivial dynamic model, such as quadrotors. \citeauthor{ma2021distributed}~\shortcite{ma2021distributed} describe DHC -- a method that efficiently utilizes the agents' communications to solve decentralized MAPF. Another communication-based learnable approach, PICO, is presented in~\cite{Li2022MultiAgentPF} and yet another in the most recent paper by~\cite{wang2023scrimp}. Overall, currently, there is a wide range of learnable decentralized MAPF solvers. In this work, we compare our method with the state-of-the-art learnable competitors and show that the former produces better quality solutions and better generalizes to the unseen maps, which is crucial for the learnable solvers.

%However, to the best of our knowledge, they all rely on the communication between the agents or on access to the global MAPF-related data (like in PRIMAL, where each agent knows the goal locations of the others). We lift these assumptions in this work.

\paragraph{MARL and HRL}

Multi-Agent Reinforcement Learning (MARL)~\cite{wong2023deep} is a separate direction in RL that specifically considers the multi-agent setting. Mainly, MARL approaches consider game environments (like Starcraft~\cite{Samvelyan2019}) in which pathfinding is not of primary importance. However, several MARL methods, such as QMIX~\cite{Rashid2018a} and MAPPO~\cite{yu2022surprising}, have been adapted specifically for the MAPF task~\cite{Skrynnik2021}. However, they rely on information sharing between the agents.

Learnable low-level policies and heuristic sub-goal allocation procedures are commonplace in many hierarchical RL (HRL) approaches tailored to single-agent problems. However, such techniques are rarely explored in MARL~\cite{Wang2022DeepRL}. Existing studies primarily demonstrate their results within simplistic environments~\cite{Tang2018HierarchicalDM}, leaving ample room for further research. Among these, PoEM~\cite{Liu2016LearningFD}, a method closely related to ours, utilizes preexisting demonstrations to identify sub-goals, which poses a significant limitation. In contrast to our approach, all the methods we know of present their findings using scenarios with a few agents.

% Much attention focuses on multi-agent learnable methods in robotics~\cite{robotics11050085}. Often, value-based approaches are used to control small groups of agents on simple maps like in~\cite{9387150} that considers a group of four agents. In \cite{8934450}, a combination of Particle Swarm Optimization and Q-Learning controlling up to 100 agents is used. \cite{8560457}, employs the model-based DynaQ method to learn agents in the knowledge exchange mode. Some works~\cite{Li_2020,app9153057} use value-based approaches with prior knowledge of how to interact with other agents. \cite{9340876} (MAPPER) uses evolutionary reinforcement learning for the MAPF task. This work also utilizes a global planner to determine sub-goals in learning one agent. In multi-agent mode, agents using ineffective policies are eliminated, and only successful agents continue to be trained.

\section{Background}

\paragraph{Multi-agent Pathfinding}
In (Classical) Multi-agent pathfinding~\cite{stern2019multi}, the timeline is discretized to the timesteps and the workspace, where $M$ agents operate, is discretized to a graph $G=(V, E)$, whose vertices correspond to the locations and the edges to the transitions between these locations. $M$ start and goal vertices are given, and each agent $i$ has to reach its goal $g_i \in V$ from the start $s_i \in V$. At each timestep, an agent can either stay in its current vertex or move to an adjacent one. An individual plan for an agent $p_i$\footnote{In MAPF literature, a plan is typically denoted with $\pi$. However, in RL, this is reserved to denote the policy. As we use both MAPF and RL approaches in this work, we denote a plan as $p$.} is a sequence of actions that transfers it between two designated vertices. The plan's cost is equal to the number of actions comprising it.
%The plan's cost is the timestep when the agent reaches the goal. 

The MAPF problem asks to find a set of $M$ plans s.t. each agent reaches the goal without colliding with other agents. Formally, two collisions are typically distinguished: a vertex collision, where the agents occupy the same vertex at the same timestep, and an edge collision, where the agents use the same edge at the same timestep.

\textit{Lifelong MAPF} (LMAPF) is a variant of MAPF where immediately after an agent reaches its goal, it is assigned to another one (via an external assignment procedure) and has to continue its operation. 

\paragraph{The Considered Decentralized LMAPF Problem}
Consider a set of agents operating in the shared environment, represented as a graph $G=(V, E)$. The timeline is discretized to timesteps $T=0,1,..., T_{max}$, where $T_{max}$ is the episode length. Each agent is located initially at the start vertex and is assigned to the current goal vertex. If it reaches the latter before the episode ends, it is immediately assigned another goal vertex. We assume that the \emph{goal assignment} unit is external to the system, and the agents' behavior does not affect the goal assignments. Each agent is allowed to perform the following actions: wait at the current vertex and move to an adjacent vertex. The duration of each action is uniform, i.e., one timestep. We assume that the outcomes of the actions are deterministic and no inaccuracies occur when executing the actions. 

Each agent has complete knowledge of the graph $G$. However, it can observe the other agents only \emph{locally}. When observing them, no communication is happening. Thus, an agent does not know the current goals or intended paths of the other agents. It only observes their locations. The observation function can be defined differently depending on the type of graph. In our experiments, we use 4-connected grids and assume that an agent observes the other agents in the area of the size $m\times m$, centered at the agent's current position.

Our task is to construct an individual policy $\pi$ for each agent, i.e., the function that takes as input a graph (global information) and (a history of) observations (local information) and outputs a distribution over actions. Equipped with such policy, an agent at each time step samples an action from the distribution suggested by $\pi$ and executes it in the environment. This continues until timestep $T_{max}$ is reached when the episode ends. Upon that, we compute the \emph{throughput} as the ratio of the number of goals achieved by all agents to episode length. We use it to compare different policies: we assert that $\pi_1$ outperforms $\pi_2$ if the throughput of the former is higher.

\paragraph{Partially Observable Markov Decision Process}

We consider a partially observable multi-agent Markov decision process~\cite{bernstein2002complexity,Pack1998}: $M=\left\langle S, A, U, P, R, O, \mathcal{O}, \gamma\right\rangle$.
At each timestep, each agent $u \in U$, where $U = {1, \dots, n}$, chooses an action $a^{(u)} \in A$, forming a joint action $\mathbf{j} \in \mathbf{J} = A^n$. This joint action leads to a (stochastic) change of the environment according to the transition function $P(s' | s, \mathbf{j}): S \times \mathbf{J} \times S \rightarrow [0, 1]$.

After that, each agent receives an individual observation $o^{(u)} \in O$ based on the global observation function $\mathcal{O}(s, a): S \times A \rightarrow O$, and an individual scalar reward $R(s, u, \mathbf{j}): S \times U \times \mathbf{J} \rightarrow \mathbb{R}$, which depends on the current state, joint action and may be different for different agents. Discount factor $0 \leq \gamma \leq 0$ determines the importance of future rewards.

To make decisions, each agent maintains an action-observation history $\tau^{(u)} \in T = (O \times A)^*$. The latter is used to condition a stochastic policy $\pi^{(u)}(a^{(u)} | \tau^{(u)}): T \times A \rightarrow [0, 1]$. The aim is to obtain (to learn) a policy $\pi^{(u)}$ for each individual agent that maximizes the expected cumulative reward over time.

\section{Learn to Follow}
\label{sec:method}

\begin{figure*}[ht!]
    \centering
    \includegraphics[width=1.0\linewidth]{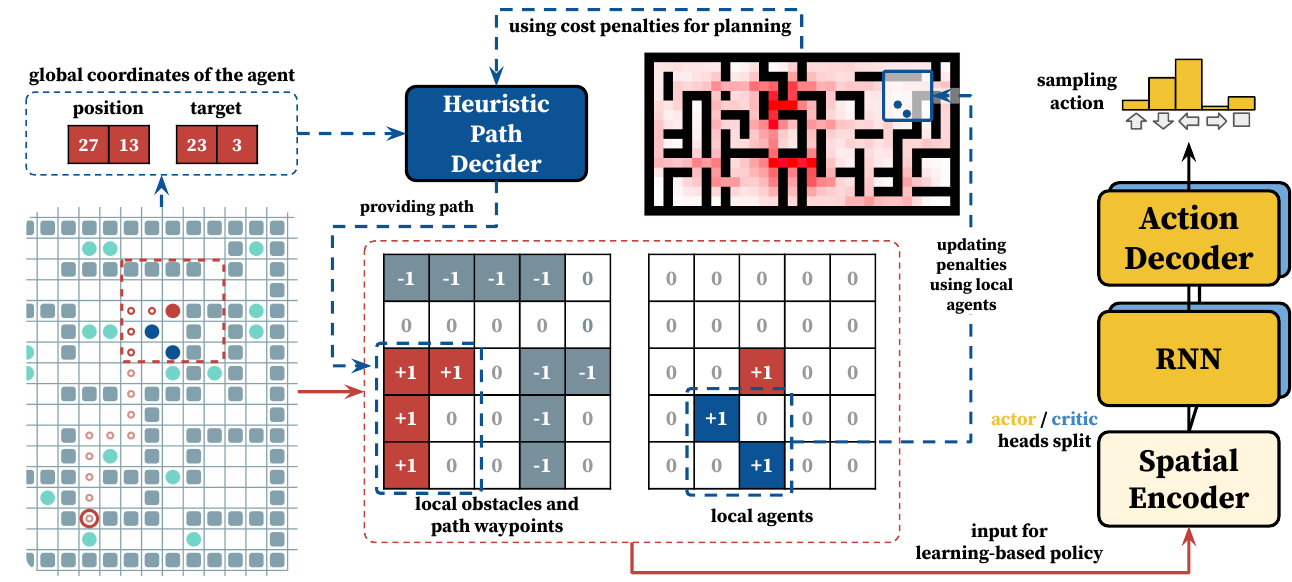}
    
    \vspace{-0.2cm}
    \caption{The general pipeline of the \textsc{Follower} approach. The action selection policy for each agent is decentralized and consists of two modules: Heuristic Path Decider, which addresses long-term path planning problem, and Learnable Follower, which addresses the short-term conflict resolution task.}
    \label{fig:main-scheme}
    \vspace{-0.2cm}
\end{figure*}

The suggested approach, which we dub \textsc{Follower}, is comprised of the two complementary modules combined into a coherent pipeline shown in Fig.~\ref{fig:main-scheme}. First, a \textit{Heuristic Path Decider} is used to construct an individual path to the goal. Then, a \textit{Learnable Follower} is invoked to reach the first waypoint on this path. This module decides which actions to take until the waypoint is reached or until the agent gets away from the path. In both cases, the sub-goal decider is called again, and the cycle repeats.

\subsection{Heuristic Path Decider}
\label{sec:method:planning}

The aim of this module is to build a path from the current location of the agent to the goal one. The static obstacles are taken into account for pathfinding, while the locations of the other agents (even the currently visible) are not; therefore, the constructed path may go through them. The rationale behind this is that the collision avoidance will be handled later on by the learnable path following policy.

A crucial design choice for this module is which individual path to build. On the one hand, A* finds the shortest (individual) path to the goal. On the other hand, when the number of agents is high and each agent is following the shortest path, congestion often arises in the bottleneck parts of the map, such as corridors or doors. This degrades the overall performance dramatically. To this end, we suggest searching not for the shortest paths but rather for the evenly dispersed paths. Intuitively, we wish to distribute the agents across the grid to decrease congestion and increase the throughput. This technique is implemented as follows.

Instead of assuming that the transition costs used by A* are uniform, we compute the individual varying transition costs associated with the cells. The individual cost of a (transition to a) cell is the sum of two components, the static and the dynamic one:

\begin{equation}
	cost(c, t) = cost_{st}(c)+cost_{dyn}(c, t).
\end{equation}

The static cost component depends on the topology of the map and does not change through the episode. The dynamic cost component, conversely, is based on the history of the observations of the agent and is dynamically updated.

To estimate the static cost of each cell, we, first, compute the average cost of the paths starting in this cell and ending in all other free cells (we use BFS algorithm for that): 

% \begin{equation}
% 	avg\_cost(c) = \sum_{c' \in V_{free}(c)}{path\_cost(c,c')/|V_{free}(c)|},
% \end{equation}
\begin{equation}
	avg\_cost(c) = \sum_{c' \in V_{free}(c)}\frac{path\_cost(c,c')}{|V_{free}(c)|},
\end{equation}
where $V_{free}(c)$ denotes the vertices reachable from $c$. 

Intuitively, the lower values of $avg\_cost(c)$ indicate that a higher number of (the shortest) paths pass through $c$, and, thus, the latter is a potential congestion attractor. Consecutively, the transition to $c$ should be penalized. This is implemented as follows: 

\begin{equation}
	cost_{st}(c) = \frac{\max_{c'\in V}(avg\_cost(c'))}{avg\_cost(c)}, 
\end{equation}

In other words, the static transition cost to a cell is $1$ only if it is the ``most rarely used'' cell of the grid, while the transition costs to the other (more frequently used) cells are higher.

The dynamic cost, $cost_{dyn}(c, t)$, is based on the personal experience of an agent and changes during the episode. It is computed as follows. 

\begin{equation}
	cost_{dyn}(c, t) = \sum_{t' \in [0,t]}{AgentAtCell(c, t')}, 
\end{equation}
where $AgentAtCell(c, t')$ is a function that returns 1 \emph{iff} some agent was observed (by the current agent) at cell $c$ at timestep $t'$ and returns 0 otherwise.

Intuitively, the dynamic cost penalizes transitions to the cells that are frequently used by the other agents. Indeed, each agent maintains its own dynamic costs. Moreover, to avoid the negative impact of over-accumulating the dynamic penalties, whenever an agent reaches its goal it resets the dynamic costs of all grid cells.

%At each timestep, the information on the locations of the locally observed agents is stored in what we call a heatmap that tells for each cell how many times the other agents were observed in this location. Further, these values are added to the transition costs and help to eliminate collisions and dead-locks between agents. Indeed, each agent maintains its own heatmap of the additional cost penalties. To avoid the negative impact of the effect of accumulating additional penalties that can affect the average true costs of the paths, whenever the agent reaches its goal and gets a new one, its heatmap of penalties is reset.

Empirically, both the precomputed transition costs and the individual dynamic costs contribute toward greater efficiency of our solver as will be shown later.

\subsection{Learnable Follower}
\label{sec:method:learning}

This module implements a learnable policy tailored to achieve the nearest waypoint of the path while avoiding a collision with the other agents. 
The policy function is approximated by a (deep) neural network and, as the agents are assumed to be homogeneous, a single network is utilized during training (a technique referred to as \textit{policy sharing}). This approach is beneficial for complex tasks and large maps where it would be infeasible to learn a separate neural network for each agent, as the number of parameters increases linearly with the number of agents. 

The input to the neural network represents the local observation of an agent and is comprised of a $2\times m\times m$ tensor, where $m$ is the observation range. The channels of the tensor encode the locations of the static obstacles combined with the current path and the other agents; see Fig.~\ref{fig:main-scheme}. 

The input goes through the \textit{Spatial Encoder} first, and then the network is split into the actor and critic heads, with the \textit{RNN blocks} designed to memorize the observation history. The output of the actor is the \textit{Action Decoder}, which produces an action distribution. The \textit{Critic Head} generates a value estimate, which is needed for training purposes only.

% The entire pipeline is trained using a policy optimization algorithm that employs a reward function, which positively rewards the agent for achieving the first waypoint of the path. 
% During the pursuit of the current waypoint, if the agent reaches it or move from it, the heuristic path decider is triggered again. This mechanism proves useful in scenarios where taking a detour to avoid congestion with other agents is more advantageous for progressing toward the global goal. 

The pipeline employs a policy optimization algorithm, rewarding the agent with \(+r\) for reaching the first waypoint.  If the agent deviates from or approaches the waypoint, the heuristic path decider is reactivated. This is advantageous in situations where taking a detour to avoid congestion with other agents is beneficial in achieving the overall goal. The focus on reaching the first waypoint provides a dense reward signal.

While the agent is rewarded for reaching the nearest waypoint, its decision-making extends beyond the immediate vicinity of that waypoint. It's important to note that the \textsc{Follower} aims to maximize rewards by navigating through multiple waypoints en route to the global goal. It takes into account potential long-term cumulative rewards, such as allowing another agent to pass and then following the path, instead of obstructing each other.

% Practicality-wise, a sub-goal is recomputed whenever the agent get $H$ cells away from it, where $H$ is a hyperparameter.
 
The task of the learning process is to optimize the shared policy $\pi^u_\theta$ (i.e. the same policy for each agent) to maximize the expected cumulative reward. During the training process, rollouts (sequences of observations, rewards, and actions) are gathered asynchronously from multiple environments with varying numbers of agents. The shared policy $\pi_\theta$ (actor network) is continually updated using the PPO clipped loss~\cite{schulman2017proximal}.

In practice, the observation history $\tau^u$ is effectively modeled using a recurrent neural network (RNN) integrated into the actor and critic heads. The actor network is parameterized by $\theta$, while the critic network is parameterized by $\phi$. In our approach, we specifically utilize the GRU architecture~\cite{chung2014empirical}.

% The introduced simplistic reward function allows to effectively train agents using relatively short rollouts. This is crucial for a lifelong setup, as here the episodes may not have a clear ending point. In our experiments, we set the rollout length to $8$.

During the decentralized inference, each agent uses a copy of the trained weights, and the other parameters remain unchanged. The proposed \textsc{Follower} scheme, despite its simplicity, allows the agent to separate the two components of the overall policy transparently and does not require the involvement of any expert data for training. Finally, the reward function used is simple and does not require involved manual shaping.

% The learning process is end-to-end, and the number of hyperparameters, that affect the result is relatively small. Finally, the reward function used is simple and does not require involved manual shaping.

\section{Experimental Evaluation}
To evaluate the efficiency of the proposed method\footnote{We are committed to open-source \textsc{Follower}.}, we conduct a set of experiments, comparing it with the state-of-the-art LMAPF algorithms on different maps. The training and evaluation of the presented approaches is held in fast and scalable  POGEMA\footnote{\dummyhref{https://github.com/AIRI-Institute/pogema}{https://github.com/AIRI-Institute/pogema}} environment. 

The learnable policy of \textsc{Follower} is implemented as the neural network of the following architecture. The \textit{Spatial Encoder} is a ResNet~\cite{he2016deep} with an additional Multi-Layer Perceptron (MLP) in the output layer. The \textit{Action Decoder} and the \textit{Critic Head} are recurrent neural networks, based on the GRU architecture~\cite{chung2014empirical}. The total number of parameters is 5M. Moreover, we additionally create a trimmed version of the network that only contains $3678$ parameters and completely omits the RNN part (see Appendix for more details). We call a solver using it \textsc{FollowerLite}. The latter is implemented purely in C++ while the regular \textsc{Follower} -- using both C++ (for pathfinding) and Python (for neural network machinery).

For training the episode length is set to $512$. The agent's field-of-view is $11 \times 11$, the number of agents varies in range: $128$, $256$. The reward for following the planned path is a small positive number, i.e. $r=0.01$. More details about the parameters of the neural network are reported in Appendix. Upon fixing the parameters, the final policy of \textsc{Follower} is trained for $1$ billion steps using a single NVIDIA A100 in approximately $18$ hours. \textsc{FollowerLite} is trained for $20$ million steps with a single NVIDIA TITAN RTX GPU in approximately $30$ minutes.

\subsection{Comparison With the Learnable Methods}
In the first series of experiments, we compare \textsc{Follower} and \textsc{FollowerLite} with the state-of-the-art learnable MAPF solvers -- \textsc{SCRIMP}~\cite{wang2023scrimp}, \textsc{PRIMAL2}~\cite{damani2021primal} and \textsc{PICO}~\cite{Li2022MultiAgentPF}. \textsc{PRIMAL2} is a seminal approach specifically tailored for solving LMAPF problems. \textsc{SCRIMP} and \textsc{PICO} are the modern decentralized MAPF solvers that were (straightforwardly) adopted by us to handle LMAPF setting. All of these solvers are indeed, decentralized, and rely on the local observations. In the experiments we utilize the conflict-handling mechanism from PRIMAL2 -- when two or more agents decide to move to the same cell, only one of them succeeds while the rest stay put. Noteworthy, \textsc{SCRIMP} has a dedicated negotiation procedure for that, which we did not modify.

%However, \textsc{PRIMAL2} assumes that the local observations contain not only the information about the current locations of the nearby agents but also about their goals on the global map. Addintionally, \textsc{PICO} and \textsc{SCRIMP} assume that the agents can communicate. The implementations of all competitors were taken from the authors' repositories. %We used the readily avalable weights (from the authors' repositories) for the neural networks of \textsc{SCRIMP} and \textsc{PRIMAL2}. \textsc{PICO} was trained by us using the authors' code.

%The code of \textsc{SCRIMP} was adopted to solve LMAPF instances.%Recall that our solver has access neither to any information about the other agents except their current locations nor to communication between the agents.

% \begin{figure}
%     \centering
%     \includegraphics[width=0.6\linewidth]{figures/01-follower-vs-primal2-mazes.pdf}
%     \caption{Average throughput on mazes. The shaded area indicates 95\% confidence intervals.}
%     \label{fig:vs-primal-mazes}
%     \vspace{-0.3cm}
% \end{figure}

\begin{figure}[htb!]
    \centering
    \includegraphics[width=0.494\linewidth]{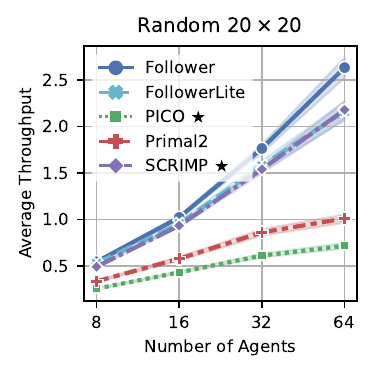}
    \includegraphics[width=0.494\linewidth]{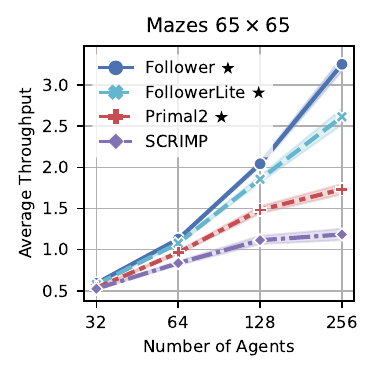}
    \vspace{-0.8cm}
    \caption{Average throughput on random and maze-like maps. 
    The shaded area indicates 95\% confidence intervals. The symbol $\star$ marks the approaches that were trained on the corresponding type of maps.}
    \label{fig:on-mazes-ranfom}
    \vspace{-0.2cm}
\end{figure}

As learnable methods assume training on a certain type of maps, we use the maps suggested by the authors of the respective baselines for a fair comparison. Specifically, we made a comparison on two types of maps -- the maze-like maps of size $65 \times 65$ on which \textsc{PRIMAL2} was originally trained, and $20 \times 20$ grids with randomly placed obstacles, that were used for training \textsc{PICO} and \textsc{SCRIMP}. The visualizations of the maps are given in Appendix. We use the readily available weights for \textsc{PRIMAL2} and \textsc{SCRIMP} neural networks from the authors' repositories. PICO was trained by us using the open-source code of its authors.  Our solvers, \textsc{Follower} and \textsc{FollowerLite}, were trained on the maze-like maps only.

For evaluation, each solver is faced with $10$ different maze-like and $40$ random maps that were not used during training. Each map is populated with an increasing number of agents, ranging from $32$ to $256$ agents for maze-like maps and from $8$ to $64$ for random ones. Start and goal locations for the agents are generated randomly. The length of the episode is set to $512$.

The results of the first series of experiments are depicted in Fig.~\ref{fig:on-mazes-ranfom}. The x-axis shows the number of agents, and the y-axis demonstrates the average throughput. Overall, on both types of maps \textsc{Follower} demonstrates the best results, notably outperforming all the competitors. The main competitor on the maze-like maps, \textsc{PRIMAL2}, shows almost twice less throughput on the instances with $256$ agents. The main competitor on random maps, \textsc{SCRIMP}, shows results equal to the lightweight version of \textsc{Follower}, i.e. \textsc{FollowerLite}. However, the results of \textsc{SCRIMP} on the maze-like maps are much worse, that indicates its low ability to generalize. \textsc{PICO} demonstrates the worst results on random maps out of all the evaluated approaches, though it was trained on this type of maps. Thus, it is excluded from the rest of the evaluations.

%The results on random (PICO) maps are presented in Fig.~\ref{fig:vs-pico-random}. Once again, both \textsc{Follower} and \textsc{FollowerLite} demonstrate superior performance across all scenarios. The poor performance of PICO can be attributed to the inherent difficulties in learning effective communication strategies for prioritizing a large number of agents. The authors of PICO trained their method on merely $8$ agents. We hypothesize that this limited population size may have impeded the acquisition of knowledge necessary for effective coordination among a larger number of agents. On the other hand, all other evaluated approaches outperform PICO, showing their ability to generalize (as they were not trained on PICO-type maps), with \textsc{Follower} being the ultimate winner.

\paragraph{Out-of-distribution evaluation.} An important quality of any learnable algorithm is its \emph{generalization}, i.e. the ability to solve problem instances that are not similar to the ones used for training. We have already seen that \textsc{Follower} generalizes well and can outperform SCRIMP on random maps though \textsc{Follower} was not trained on this type of maps. Now we run an additional test where we compare \textsc{Follower}, \textsc{FollowerLite}, \textsc{PRIMAL2} and \textsc{SCRIMP} on two (unseen during learning) maps from the well-known in the MAPF community MovingAI benchmark~\cite{stern2019multi}: \texttt{den520d} and \texttt{Paris\_1}. The former map was taken from a video game, while the latter one correspond to the part of a real city. Their topologies are quite different from the one of the maps used for training. Both of the maps were downscaled to the size of $64 \times 64$. 

\begin{figure}
    \centering
    \includegraphics[width=0.49\linewidth]{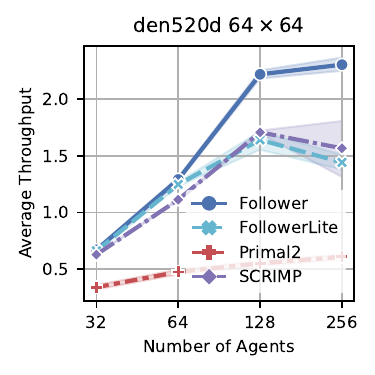}
    \includegraphics[width=0.49\linewidth]{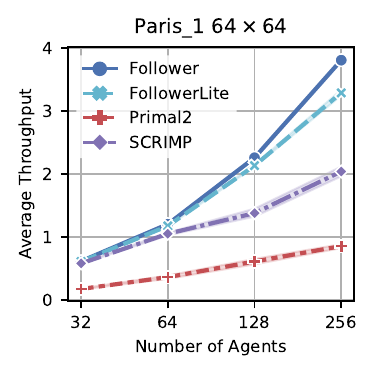}
    \vspace{-0.3cm}
    \caption{The results on MovingAI \texttt{den520d} and \texttt{Paris\_1} maps. The shaded area indicates 95\% confidence intervals.}
    \label{fig:vs-primal-out}
    \vspace{-0.4cm}
\end{figure}

The results of these experiments are presented in Fig.~\ref{fig:vs-primal-out}. Again \textsc{Follower} significantly outperforms all the competitors. \textsc{PRIMAL2} demonstrates very low throughput on both maps, that indicates its bad ability of generalization. Compared to \textsc{PRIMAL2}, \textsc{SCRIMP} shows itself much better in terms of generalization, but in the best case it is only able to demonstrate results comparable to the lightweight version of the proposed approach, i.e. \textsc{FollowerLite}. Additional results of the out-of-distribution experiments are presented in Appendix.

\subsection{Comparison With Non-Learnable Approaches}
\label{sec:evaluation:vs-rhcr}
We compare \textsc{Follower} and \textsc{FollowerLite} with two non-learnable approaches -- \textsc{RHCR}\footnote{https://github.com/Jiaoyang-Li/RHCR}~\cite{li2021lifelong} and \textsc{PIBT}\footnote{https://github.com/Kei18/pibt2}~\cite{okumura2022priority}.  These two approaches are in two different polarities -- \textsc{RHCR} is the state-of-the-art search-based approach aiming at the high-quality (i.e. high-throughput) LMAPF solutions at the expense of the limited scalability, while \textsc{PIBT} is the state-of-the-art rule-based approach that is extremely fast, but provides solutions of a moderate quality.   %In contrast to the proposed method, RHCR is a centralized approach that coordinates all the agents and does not restrict the observation and/or communication abilities of the agents. 

\textsc{RHCR} solver requires setting a time limit for planning. We set it either to $1$ or $10$ seconds (both variants are reported with the according names). We chose PBS~\cite{ma2019searching} as the planning approach since it showed the best results in the original paper.  We have also tuned the planning horizon $(2, 5, 10, 20)$, the re-planning rate $(1, 5)$ and found that the best throughput is achieved by RHCR when the first parameter is set to $20$ and the second one to $5$ (see Appendix for more details). We use these values in our experiments. The other parameters of RHCR are left default. 

The comparison is performed on the \texttt{warehouse} map from the original RHCR paper~\cite{li2021lifelong}. The maximum number of agents is limited to $192$, according to restrictions for starting locations introduced in \cite{li2021lifelong}. We generated $10$ random instances for each number of agents for evaluation.

\begin{figure}[htb!]
    \vspace{-0.3cm}
    \centering
    \includegraphics[width=0.494\linewidth]{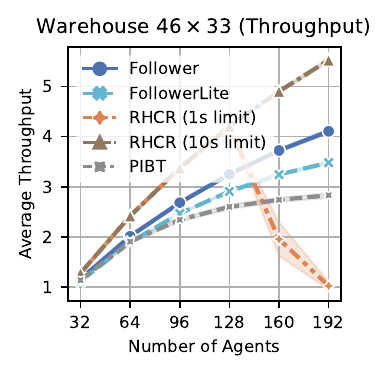}
    \includegraphics[width=0.494\linewidth]{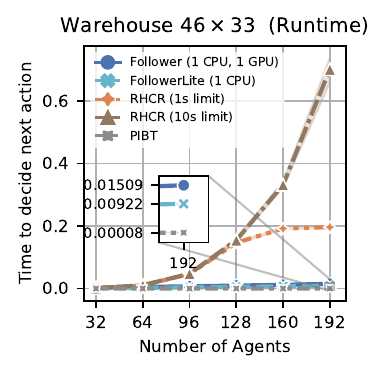}
    \vspace{-0.8cm}
    \caption{Average throughput and runtime on \texttt{warehouse} map. The shaded area indicates 95\% confidence intervals.}
    \vspace{-0.2cm}
    \label{fig:vs-rhcr}
    
\end{figure}

% The results are presented in Fig.~\ref{fig:vs-rhcr}. As can be seen in Fig.~\ref{fig:vs-rhcr}~(Throughput) both versions of \textsc{RHCR} significantly outperform our solvers when the number of agents is up to $128$. However, when this number increases to $160$ and $192$ the performance of \textsc{RHCR} with 1s time cap degrades dramatically and it gets outperformed by both \textsc{Follower} and \textsc{FollowerLite}. This pinpoints the principal limitation of the centralized approach: it does not scale well to large number of agents when a strict time limit for finding a MAPF solution is imposed. \textsc{PIBT} is able to outperform neither \textsc{Follower} nor \textsc{FollowerLite} in terms of average throughput.

The results are presented in Fig.~\ref{fig:vs-rhcr}. As can be seen on the left part of the Fig.~\ref{fig:vs-rhcr}, both versions of \textsc{RHCR} significantly outperform the other solvers when the number of agents is up to $128$. However, when this number increases to $160$ and $192$, the performance of \textsc{RHCR} with a 1s time cap degrades dramatically. It is then outperformed by both \textsc{Follower} and \textsc{FollowerLite}. This highlights the principal limitation of the centralized approach: it does not scale well to a large number of agents, especially when a strict time limit for finding a MAPF solution is imposed. \textsc{PIBT} does not have such a problem with scalability but its throughput is the lowest among all the evaluated methods.

To better understand how the runtime of the evaluated methods is affected by the increasing number of agents, see the right pane of the Fig.~\ref{fig:vs-rhcr}. In this plot each data point indicates how much time on average is spent to decide the next action. Indeed, \textsc{Follower} needs much less time to choose an action and, consequtively, scales better to the increasing number of agents compared to \textsc{RHCR}. 
% Moreover, in practice, it can be parallelized (see \textsc{FollowerLite} (32 CPUs)), that can decrease the time required to decide next actions for all the agents even further. 
As expected, \textsc{PIBT} is the fastest approach, as its rule-based procedures are computationally cheap compared to the ones used by \textsc{RHCR} and \textsc{Follower}. Recall, however that the throughput of \textsc{PIBT} is inferior. Moreover, in practice our method can be effectively parallelized (to run individually on each agent) and this will (as we expect) lower down its runtime further on.

%which expectedly will lower down its runtime.

%but \textsc{PIBT} cannot be efficiently parallelized and provides solutions of the worst quality among the compared solvers.

\subsection{Ablations}

\begin{figure}[htb!]
    \vspace{-0.1cm}
    \centering
    \includegraphics[width=0.494\linewidth]{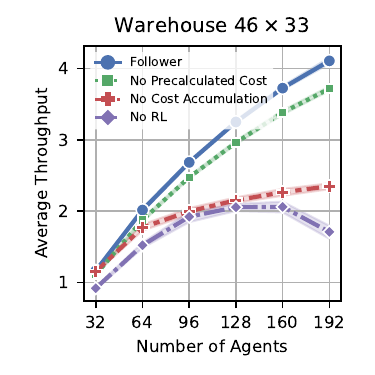}
    \includegraphics[width=0.494\linewidth]{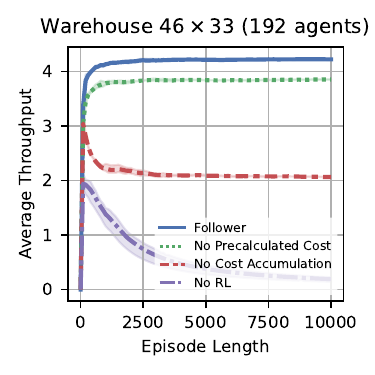}
    \vspace{-0.8cm}
    \caption{Impact of \textsc{Follower} components on it's performance. The shaded area indicates 95\% confidence intervals.}
    \label{fig:ablations}
    \vspace{-0.4cm}
\end{figure}

In this experiment, we investigate the impact of different components on the performance of \textsc{Follower}. To this end we turn them off and run the resultant solver on the \texttt{warehouse} map from the RHCR paper. Specifically, \textsc{Follower} (no RL) omits the learnable policy. At each timestep, it plans a path, taking into account other observable agents as static obstacles, and selects its first action for the execution. If no path is found a random action is picked. \textsc{Follower} (no cost accumulation) and \textsc{Follower} (no precalculated cost) use both planning and learning components, but they do not utilize one of the introduced techniques that penalize transitions to the frequently used cells.

%as described in Section~\ref{sec:method:planning}. 

The results are shown in Fig.~\ref{fig:ablations} (left pane). First, note that the performance of \textsc{Follower} (no RL) is inferior, which justifies the importance of the learnable policy and its contribution to the efficiency of \textsc{Follower}. The same can be said about the cost-penalizing techniques. Overall, it is clear that \emph{all} components of \textsc{Follower} are crucial; omitting any of them results in a notable degradation of performance.

In addition, we run \textsc{Follower} with different episode lengths (up to $10,000$), as the initial distribution of the agents can be very different from the distribution that happens after some time. The results are shown in Fig.~\ref{fig:ablations} (right pane). Notably, the absense of RL policy and dynamic cost accumulation lead to a very low throughput in the limit. We explain this by the congestion that occurs and grows like a rolling snowball and prevent the agents from reaching their goals.  Indeed, \textsc{Follower} copes well with this as its throughput monotonically increases with the episode length and then plateaus.

\subsection{Summary}

The proposed approach surpasses learnable decentralized competitors, especially when the number of agents is large or when dealing with maps different from the training ones. Both the learnable component and the cost-penalizing techniques are essential to \textsc{Follower}'s performance. Furthermore, \textsc{Follower} scales much better than the state-of-the-art search-based solver and provides solutions of the better quality compared to modern rule-based solver.

\section{Conclusion}
\label{sec:conclusion}
In this study, we addressed the challenging problem of decentralized lifelong multi-agent pathfinding. We proposed a solution that leverages heuristic search for long-term planning and reinforcement learning for short-term conflict resolution. Our method consistently outperforms decentralized learnable competitors. Moreover, it provides the better trade-off between the scalability and the solution quality compared to the modern search-based and rule-based planners. Directions for future research may include: enriching the action space of the agents, handling uncertain observations and external stochastic events.

\bibliography{aaai24}

\clearpage

\section{Appendix}

\subsection{Source Code}

The source code related to this paper will be made publicly available upon publication. Until then, we invite individuals interested in accessing the code to contact us directly via email.

% \subsection{Anonymized Code}
% The anonymized code of \textsc{Follower} is available at \dummyhref{https://anonymous.4open.science/r/learn-to-follow}{https://anonymous.4open.science/r/learn-to-follow}.

% The repository includes all resources required to replicate the results of \textsc{Follower} and \textsc{FollowerLite}, including training and evaluation scripts, pre-trained weights, and the dataset used for training and evaluation. 
% In this appendix, we will reference to the specific maps from the dataset using the names provided in the repository.

\subsection{Limitations}
\label{sec:imitations}
Similarly to a wide range of other works focusing on learnable (L)MAPF solvers (including the ones we compare with), we rely on the following assumptions. The map of the environment is accurate and does not change. The agents have perfect localization and mapping abilities and execute actions accurately (and their moves are synchronized). All these may be considered as the limitations, as in real world, e.g. in robotic applications, many of the assumptions do not hold.

Another limitation is that the suggested approach, as any other (known to us) learnable (L)MAPF solver, is not able to provide theoretical guarantees of completeness/optimality.

\subsection{The Effect Of Penalizing The Transition Costs}\label{appendix:penalties}

As explained in the main body of the paper, for pathfinding with A* each agent computes an individual transition cost to each cell of the grid that is updated during the episode. Each time some other agent is observed at a certain cell its transition cost is increased. This helps to avoid areas that are often used by many agents and thus to pro-actively avoid congestion, as shown in the main body of the paper.

\begin{figure}[htb!]
    \centering
    \captionsetup{justification=centering}
    \begin{subfigure}[b]{0.4\linewidth}
        \centering
        \includegraphics[width=\linewidth]{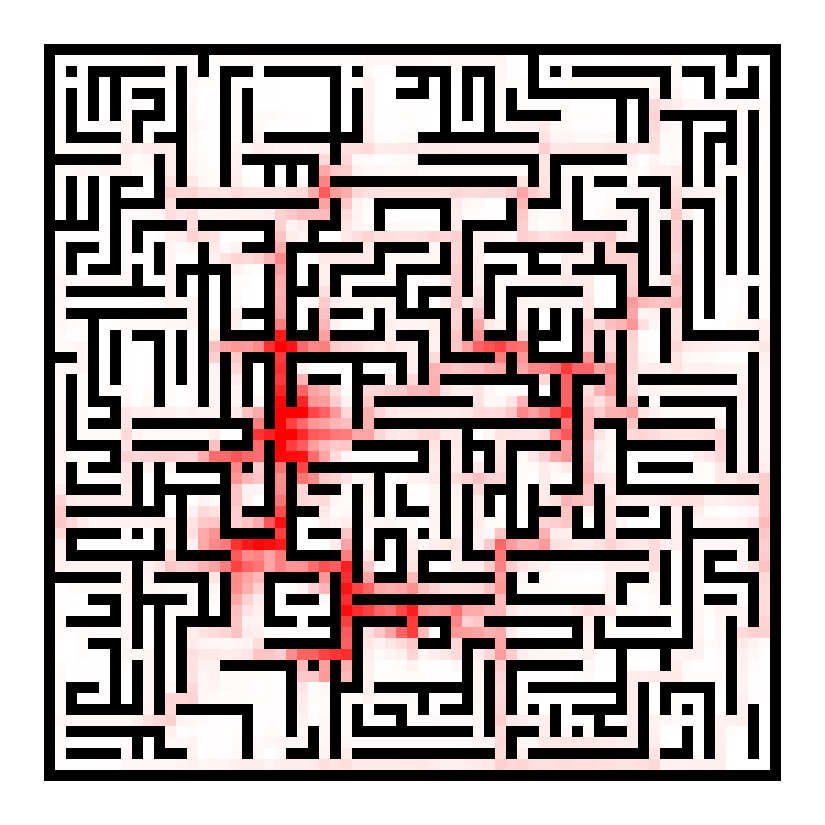}\
        \vspace{-20px}
        \caption{\textsc{Follower}\\(uniform transition costs)}
    \end{subfigure}
    \begin{subfigure}[b]{0.4\linewidth}
        \centering
        \includegraphics[width=\linewidth]{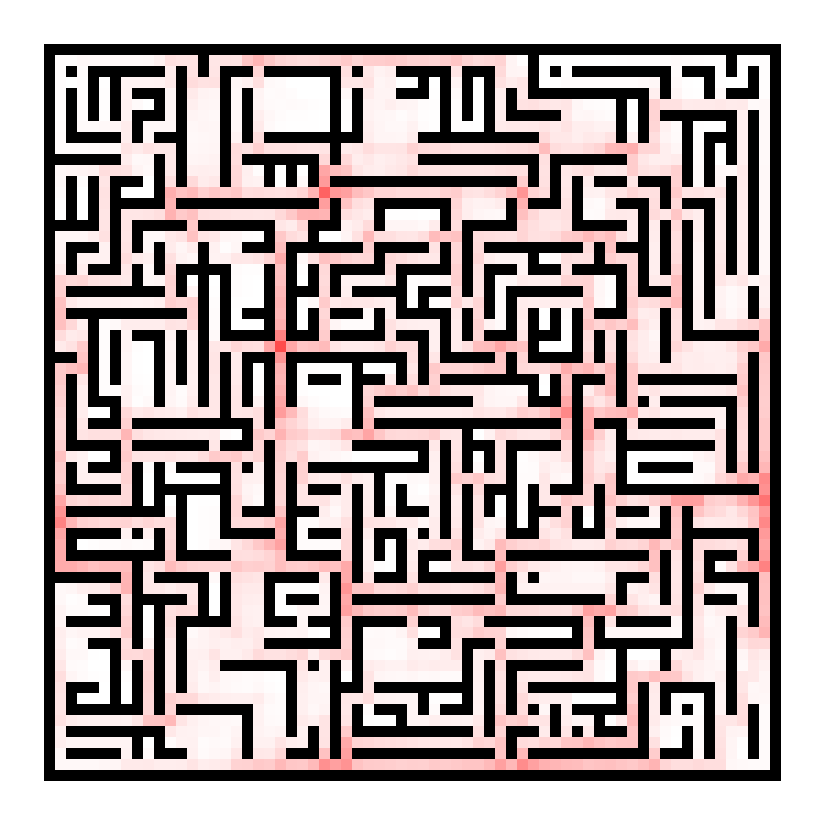}\
        \vspace{-20px}
        \caption{\textsc{Follower}\\(varying transition costs)}
    \end{subfigure}
    \begin{subfigure}[b]{0.4\linewidth}
        \centering
        \includegraphics[width=\linewidth]{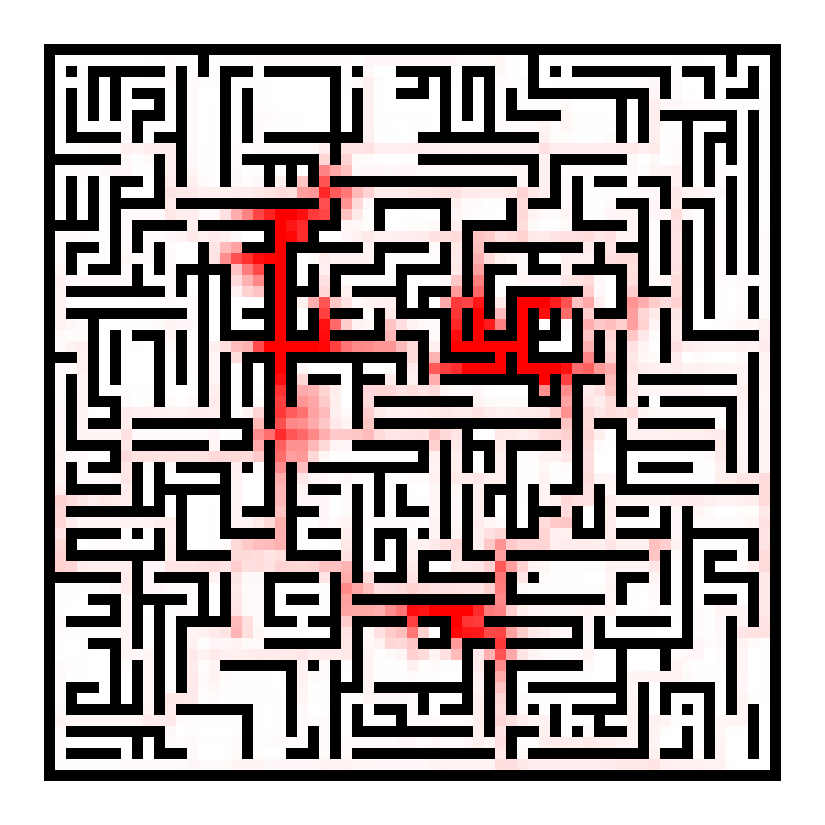}\
        \vspace{-20px}
        \caption{\textsc{FollowerLite} (uniform transition costs)}
    \end{subfigure}
    \begin{subfigure}[b]{0.4\linewidth}
        \centering
        \includegraphics[width=\linewidth]{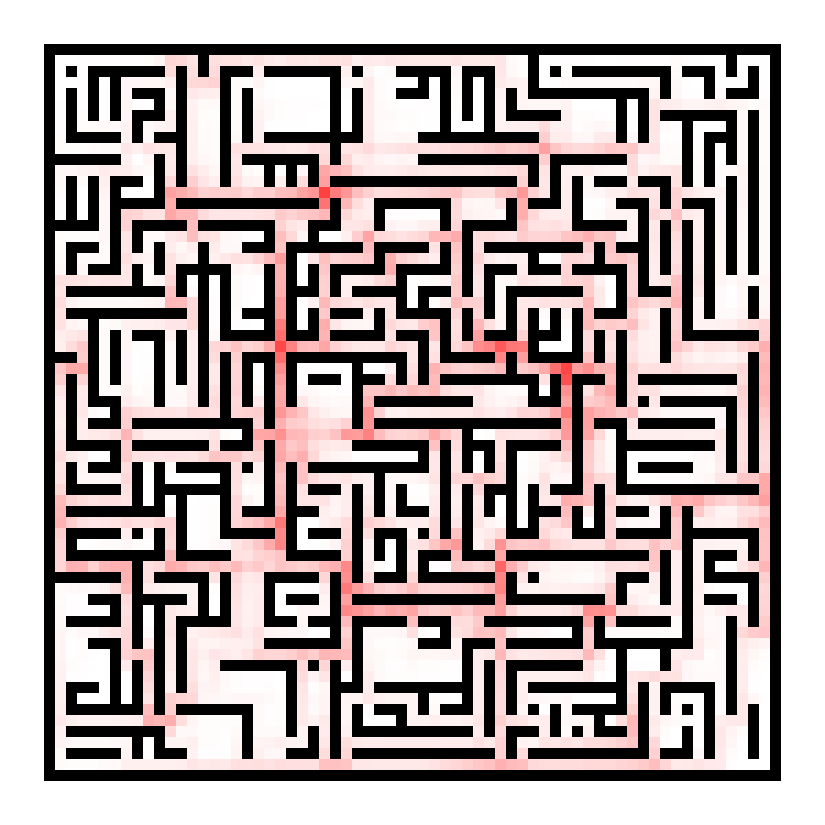}
        \vspace{-20px}
        \caption{\textsc{FollowerLite} (varying transition costs)}
    \end{subfigure}
    \caption{Heatmaps representations of how often the agents visited certain areas when solving a particular LMAPF instance containing 128 agents. The more intense red collor is -- the more frequently the corresponding area has been visited.
    }\label{fig:transition_cost}
\end{figure}

Here we wish to demonstrate the effect of the introduced cost penalizing technique visually. In Fig.~\ref{fig:transition_cost} one of the problem instances from our dataset in shown. The left part of the figure shows the result of solving this instance (that contained 128 simultaneously moving agents for 512 timesteps) by \textsc{Follower} and \textsc{FollowerLite} that omit the cost penalizing technique (i.e.  all the cells have transition cost 1 and it never changes). The more intense the (red) color is -- the more frequently the corresponding cell was visited by the agents. Clearly, in this case the agents tend to use the central part of the map frequently. This makes it hard for the agents to avoid each other in the narrow passages that are plentiful on this map. On the other hand, when the individual varying transition costs are employed (right part of Fig.~\ref{fig:transition_cost}), the agents get evenly distributed across the map, which prevents congestion and increase the performance (as confirmed by the experiments, reported in the main body of the paper).

\subsection{Additional Out-Of-Distribution Evaluation}
In the main body of the paper we have already reported the results obtained on the maps that are not alike the ones used for training the considered learnable solvers. Here we provide the additional results of \textsc{Follower}, \textsc{FollowerLite}, PRIMAL2 and SCRIMP on $4$ additional out-of-distribution maps taken from the MovingAI benchmark: \texttt{Boston\_0}, \texttt{den312d}, \texttt{room-64-64-16} and \texttt{room-64-64-8}.

\begin{figure*}
    \centering
    \includegraphics[width=\textwidth]{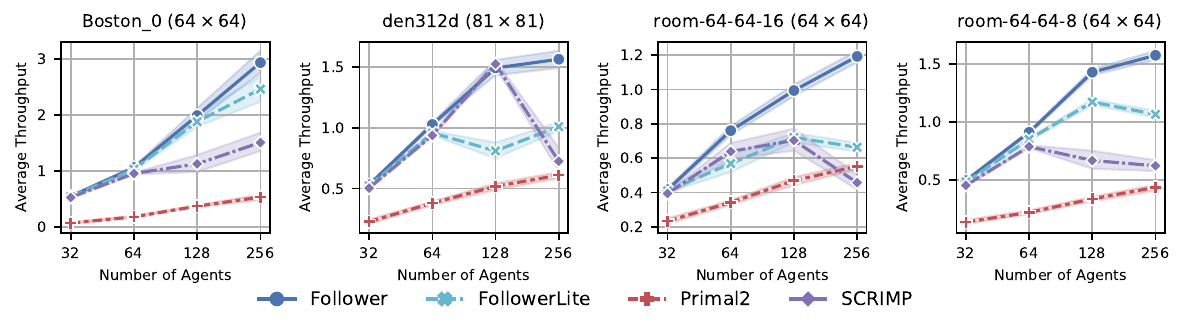}
    \caption{The results of the additional evaluation of all learnable solvers on out-of-distribution maps, i.e. 4 maps taken from the well-known MovingAI MAPF benchmark. The shaded area indicates 95\% confidence intervals.}
    \label{fig:ood-results}
    \vspace{-0.4cm}
\end{figure*}

The results presented in Fig.~\ref{fig:ood-results} confirm the claims made in the main text of the paper. First, PRIMAL2 is the worst solver in terms of generalization and ability to achieve high troughput on the maps that are not alike the ones it was trained for. Second, neither SCRIMP nor PRIMAL2 are able to compete with \textsc{Follower} when the number/density of agents is high. Third, the throughput of \textsc{FollowerLite}  is lower than the one of \textsc{Follower}, which is expected as the neural network of the former has much less parameters (3K vs. 5M) and does not contain the RNN (GRU) blocks. 

\subsection{Tuning RHCR Parameters}\label{appendix:rhcr-tuning}

As it was mentioned in the main text, we have tuned the parameters of RHCR before conducting the empirical comparison to our method. We varied the values of such RHCR parameters as planning horizon (2, 5, 10, 20), re-planning rate (1,5) and time limit for each re-planning attempt (1s, 10s). Planning horizon parameter controls for how many time steps the resultant plans will be collision-free. E.g. when it equals 10 it is guaranteed that for the next 10 time steps the agents following the constructed plans will not collide. Re-planning rate determines how frequently (in time steps) reconstruction of the plans (for all agents) occurs. Time limit parameter restricts the amount of time (in seconds) which is alotted for each re-planning attempt. %In case of exceeding the given threshold RHCR stops the run of ``main'' solver, which is PBS, and runs a faster LRA* algorithm that may output plans with collisions.
\begin{figure}[htb!]
    \centering
     \includegraphics[width=0.494\linewidth]{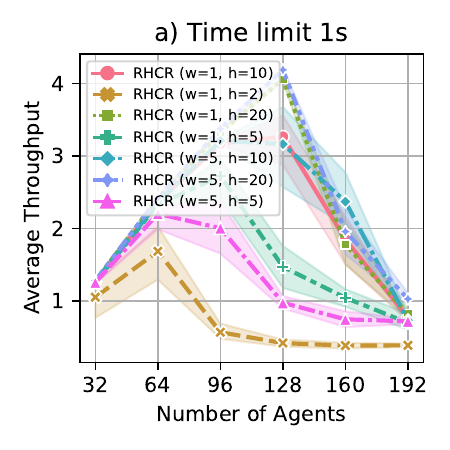}
    \includegraphics[width=0.494\linewidth]{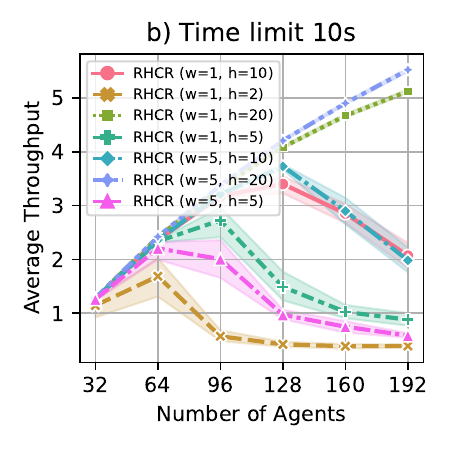}
    \caption{Evaluation of RHCR with different values of its user-specified parameters: $w$ -- re-planning rate, $h$ -- planning horizon.}
    \label{fig:rhcr-tune}
\end{figure}

Fig.~\ref{fig:rhcr-tune} demonstrates the results of different versions of RHCR (note that planning horizon cannot be lower than re-planning rate). The best average throughput is achieved by RHCR with re-planning rate 5 and planning horizon 20, denoted as ($w=5$, $h=20$) in the figure. Noteworty, the same values of these parameters were used for the experimental evaluation on the \texttt{warehouse} map in the original RHCR paper. Thus, the results of this version are included into the main part of our paper.

\subsection{Additional Training Details}
    \begin{figure*}[ht!]
    \centering
    \includegraphics[width=0.27\linewidth]{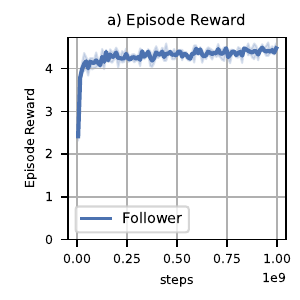}
    \includegraphics[width=0.27\linewidth]{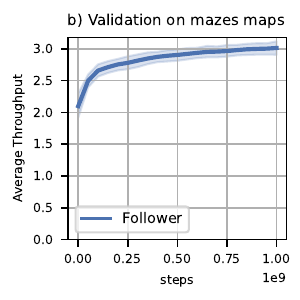}
    \includegraphics[width=0.27\linewidth]{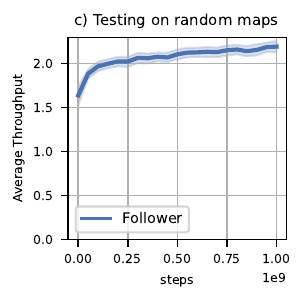}
    
    \caption{Training progress \textsc{Follower}. (a) shows the average running episode reward over the course of training (recall that an agent receives a small positive reward, i.e. $0.01$, each time it achieves a waypoint on the path to the current goal). (b) shows the validation results on a subset of training maps (i.e. maze maps) with the fixed seeds and $256$ agents. Additionally, we tested the checkpoints on random maps with $20\%$ density and $64$ agents to assess the algorithm's generalization ability during training. The results are shown in (c). All the reported indicators are averaged over three runs, and the shaded area corresponds to the 95\% confidence intervals.}
    \label{fig:training-follower}
    \end{figure*}

    As reported in the main body of the paper, after tuning and fixing the hyper-parameters required by the learnable component of \textsc{Follower}, the latter was trained for one billion steps on A100 GPU ($18$ hours of training). Only maze-like maps were used for training. Fig.~\ref{fig:training-follower} illustrates how the performance changed while training.

    Fig.~\ref{fig:training-follower} (a) shows the reward plot. Some perturbations of the reward are noticeable, which can be explained by the random sampling of maps and problem instances, among which both simple and complex setups are present. Overall, the reward consistently grows (and then stabilizes) which means that learning is, indeed, happening.

    During training we also regularly saved the current weights of the neural network and invoked \textsc{Follower} on the subset of the training maps with the fixed seeds and $256$ agents. The results are shown in Fig.~\ref{fig:training-follower}~(b). Clearly, the throughput is increasing, indicating that a single agent is not only learning to follow its path effectively but also acquiring the skills needed to cooperatively solve non-trivial LMAPF instances. 

    Moreover, during learning, we tested  \textsc{Follower} on random maps that were not part of the training dataset. The results are shown in Fig.~\ref{fig:training-follower} (c). Again, we observed an increase in throughput, indicating that our solver does not overfit to the training maps but is capable of adapting to and solving arbitrary LMAPF problems effectively.

\begin{figure*}[t!]
    \centering
    \captionsetup{justification=centering}
     \begin{subfigure}[b]{0.19\textwidth}
         \centering
         \includegraphics[height=3cm]{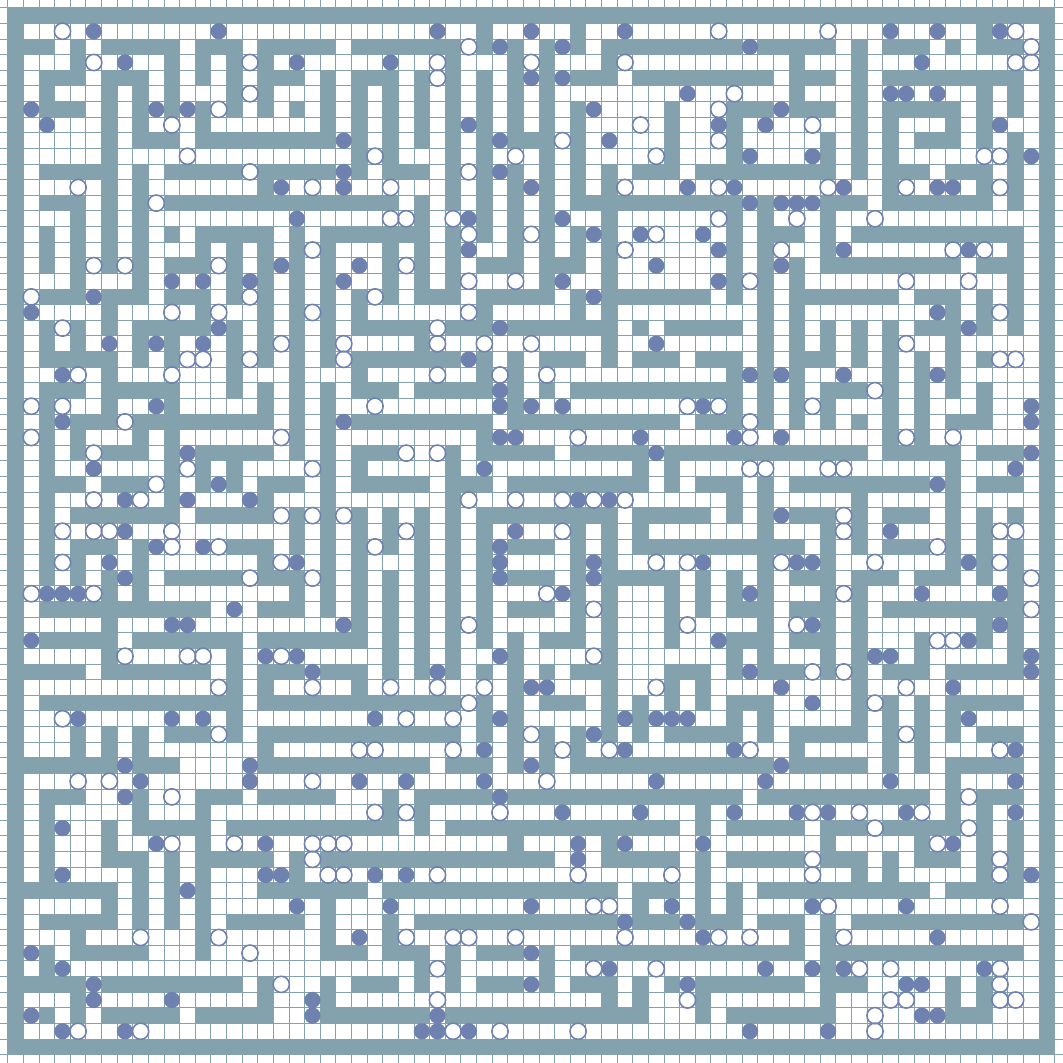}
         \caption{Mazes-wc3-od70\\$65\times65$}
         \label{fig:maze_map}
    \end{subfigure}
    \begin{subfigure}[b]{0.19\textwidth}
         \centering
         \includegraphics[height=3cm]{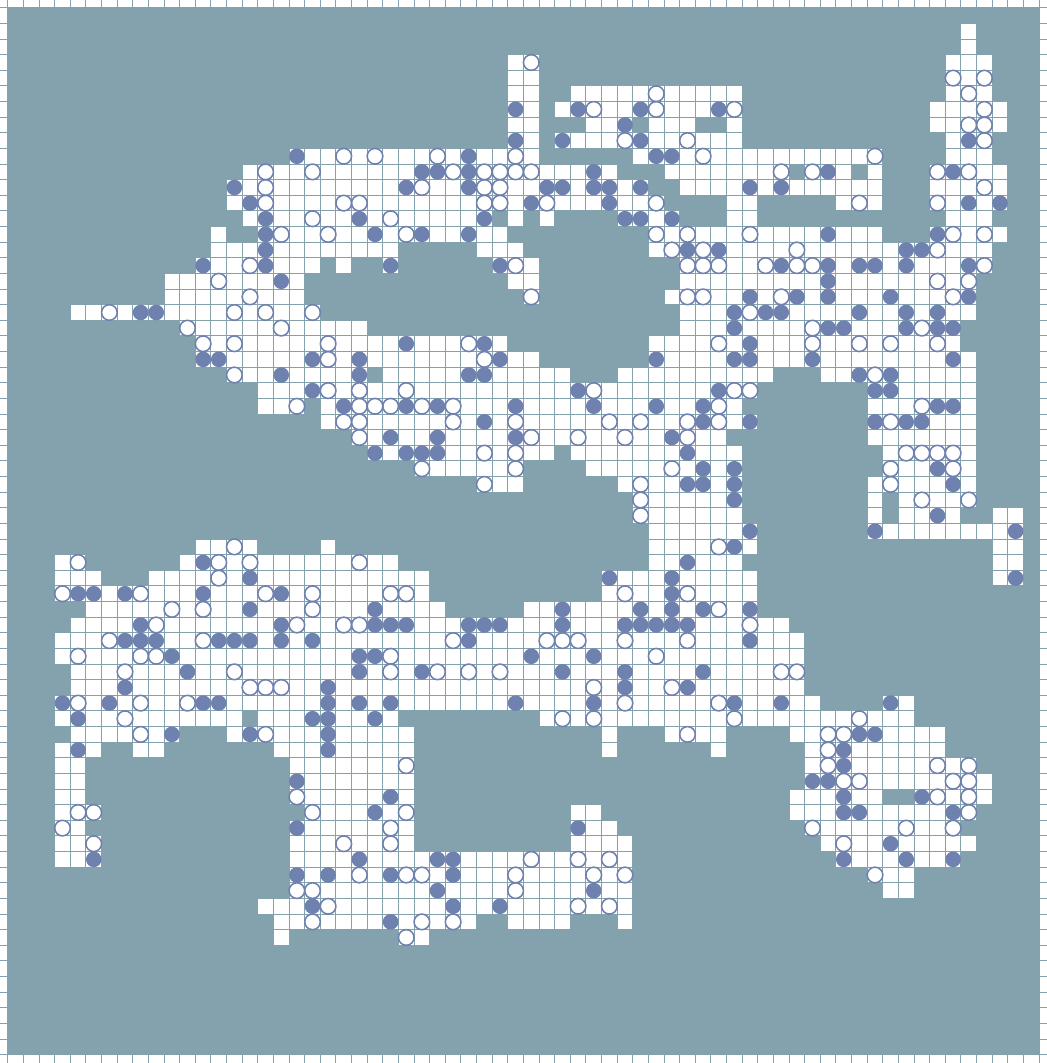}
         \caption{Den520d\\$64\times64$}
    \end{subfigure}
    \begin{subfigure}[b]{0.19\textwidth}
         \centering
         \includegraphics[height=3cm]{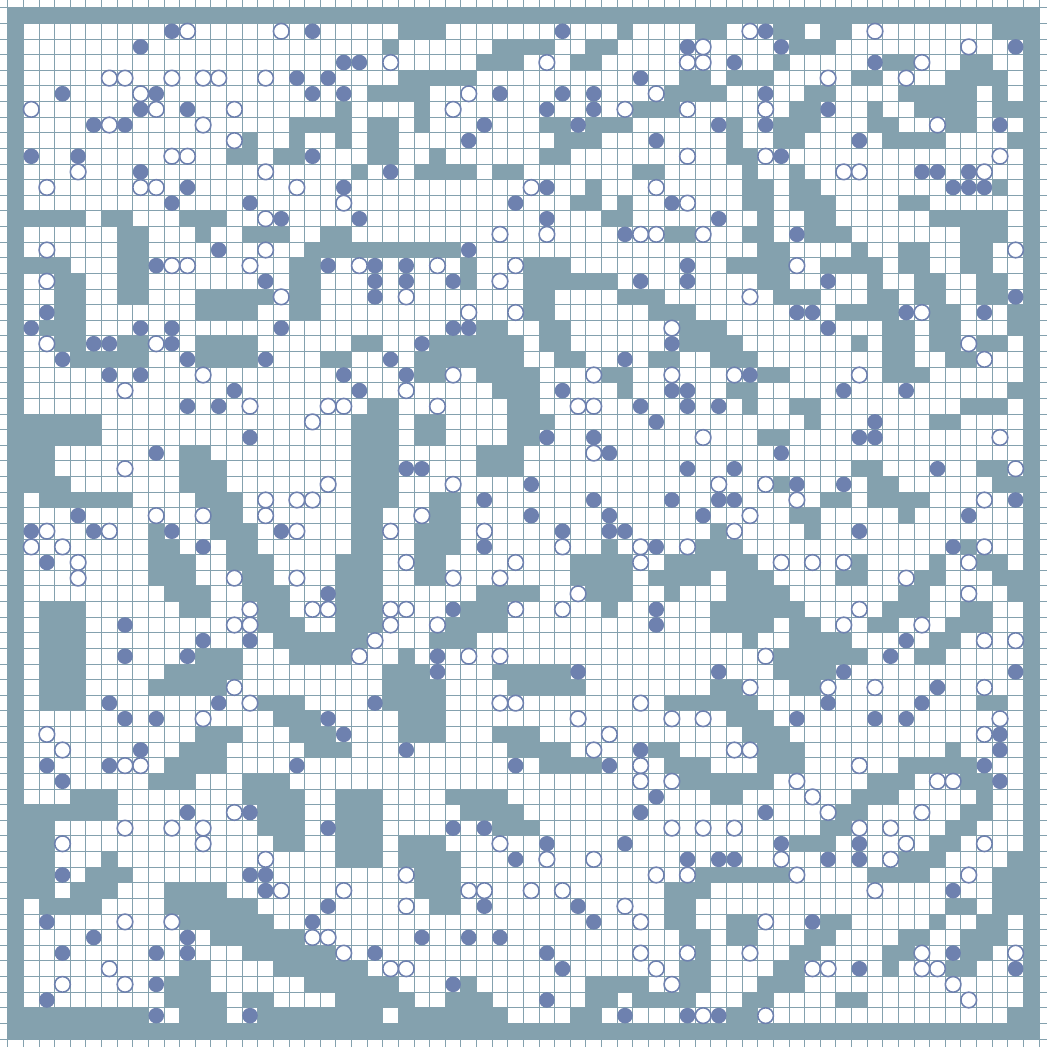}
         \caption{Paris\_1 \\$64\times64$}
    \end{subfigure}
    \begin{subfigure}[b]{0.19\textwidth}
         \centering
         \includegraphics[height=3cm]{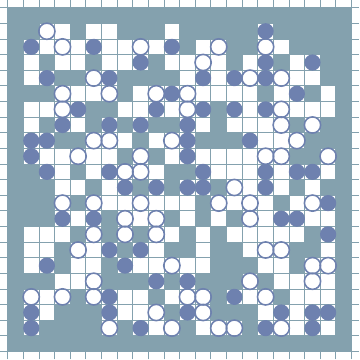}
         \caption{PICO-s21-od30\\$20\times20$}
    \end{subfigure}
    \begin{subfigure}[b]{0.19\textwidth}
         \centering
         \includegraphics[height=3cm]{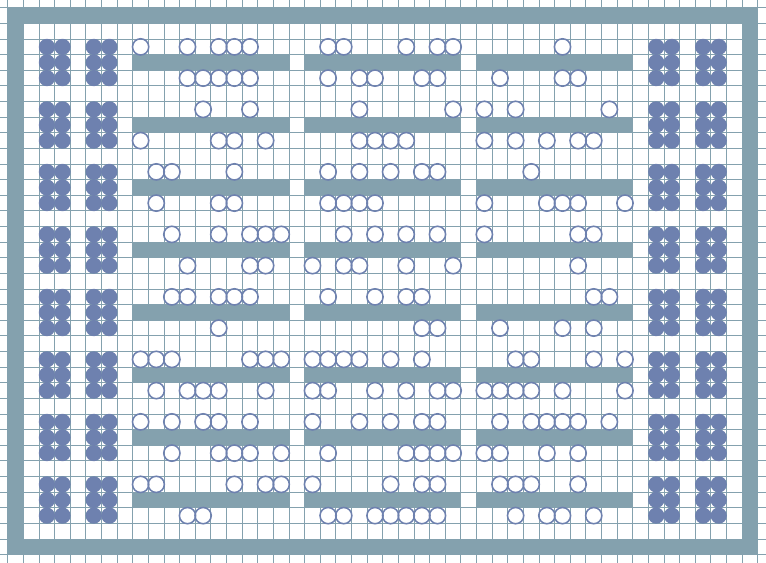}
         \caption{Fulfillment warehouse \\$46\times33$}
    \end{subfigure}

    \caption{Visualized examples of the maps with a maximum number of agents used for empirical evaluation.
    Initial locations of the agents are represented by the filled circles, while their (first) goals are represented by the empty ones. 
    }
    \label{fig:map_examples}
    % \vspace{-0.3cm}
\end{figure*}

    \begin{table}[t!]

        \centering
        \caption{The hyperparameters of \textsc{Follower} and \textsc{FollowerLite}. The \emph{tune range} column indicates the range for parameters adjusted through a hyperparameter optimization procedure.
        }
        % \vspace{0.28cm}
        \small
        \label{table:parameters}
        \resizebox{\linewidth}{!}
        {
        \begin{tabular}{l|lr|lr}

            \toprule
            Hyperparameter                & FollowerLite    & Tune range                       & Follower      & Tune range           \\
            \midrule
            \midrule
            Adam learning rate            & $0.000133$      & $0.0001$ -- $0.0002$             & $0.00022$     & $0.0001$ -- $0.0002$ \\
            $\gamma$ (discount factor)    & $0.971$         & $0.95$ -- $0.99$                 & $0.976$       & $0.95$ -- $0.99$     \\
            Recurrence rollout            & -               & -                                & $8$           & $[4, 8, 16, 32]$     \\
            Clip ratio                    & $0.2$           & -                                & $0.2$         & -                    \\
            Batch size                    & \num{16384}     & -                                & \num{16384}   & -                    \\
            Optimization epochs           & $1$             & -                                & $1$           & $[1, 5, 10]$         \\
            Entropy coefficient           & $0.0157$        & $0.01$ -- $0.05$                 & $0.023$       & $0.01$ -- $0.05$     \\
            Value loss coefficient        & $0.5$           & -                                & $0.5$         & -                    \\
            GAE$_\lambda$                 & $0.95$          & -                                & $0.95$        & -                    \\
            \midrule
            MLP hidden size               & $16$            & $[8, 16, 32]$                    & $512$         & $[256, 512]$         \\
            ResNet residual blocks        & $1$             & $[1, 2]$                         & $8$           & $[2, 4, 6, 8, 10]$   \\
            ResNet filters                & $8$             & $[8, 16, 32]$                    & $64$          & $[32, 64, 128]$      \\
            GRU hidden size               & -               & -                                & $256$         & $[256, 512]$         \\
            Activation function           & ReLU            & -                                & ReLU          & -                    \\
            Network Initialization        & orthogonal      & -                                & orthogonal    & -                    \\
            Number of agents              & $[128, 256]$    & -                                & $[128, 256]$  & -                    \\
            Rollout workers               & $4$             & -                                & $8$           & -                    \\
            Envs per worker       & $4$             & -                                & $4$           & -                    \\
            Training steps                & $2 \times 10^7$ & -                                & $10^9$        & -                    \\
            \midrule
            Observation radius            & $5$             & -                                & $5$           & -                    \\
            Observation patch             & $11\times11$    & -                                & $11\times11$  & -                    \\
            Network input size            & $7\times7$      & $[3, 5, 7, 9]$                   & $11\times11$  & -                    \\
            Network parameters            & \num{3678}      & -                                & \num{5150406} & -                    \\
            \bottomrule
        \end{tabular}
        }
    \end{table}

\subsection{Hyperparameters}\label{appendix:hyperparameters}

    Table~\ref{table:parameters} presents the hyperparameters of \textsc{Follower} and \textsc{FollowerLite}. 

    The hyperparameters for which the tuning range is given (e.g. learning rate, GRU hidden size, etc.) are optimized using Bayesian search. 

    The observation radius has been set to $11\times11$ as it is commonly used in similar learning-based methods (with whom we compare). The parameters for the number of \emph{rollout workers}, \emph{environments per worker}, and \emph{training steps} are empirically determined to decrease the overall learning time of the algorithm. For the remaining paramaters (value loss coefficient, GAE$_\lambda$, activation function, network initialization), we have used the default values provided in the SampleFactory framework\footnote{\dummyhref{https://github.com/alex-petrenko/sample-factory}{https://github.com/alex-petrenko/sample-factory}}.

    We conducted a hyperparameter sweep with approximately $150$ runs, amounting to around $200$ GPU hours. The models for \textsc{Follower} were trained for $100$ million steps. Our optimization target was the average throughput on the validation set, which consisted of $15$ mazes from the training set with fixed seeds.

\subsection{Maps Visualizations}\label{appendix:maps}

Fig.~\ref{fig:map_examples} illustrates examples of the maps used for testing. The presented names of the maps correspond to the names in our repository.

Initial positions of the agents are shown by the filled circles, while their (initial) goals are represented by the empty ones. Each agent is assigned a unique goal initially. Subsequent LMAPF goals are randomly generated, ensuring that a feasible path from the agent's current location to the goal exists. The goals for each agent are generated independently using a fixed seed, ensuring consistency and enabling fair testing of the algorithms (i.e. each algorithm gets the same start/goals locations).

\end{document}